# Multi-Resolution Feature Fusion U-Net for Magnetic Resonance Imaging Segmentation

Eirini Cholopoulou
*Department of Computer Science and Biomedical*
*University of Thessaly*
Lamia, Greece
echolopoulou@uth.gr

Dimitrios E. Diamantis
*Department of Computer Science and Biomedical*
*University of Thessaly*
Lamia, Greece
didiamantis@uth.gr

Dimitris K. Iakovidis
*Department of Computer Science and Biomedical*
*University of Thessaly*
Lamia, Greece
diakovidis@uth.gr

***Abstract*—The segmentation of anatomical structures in medical images and particularly in MRI scans, is essential for clinical diagnosis and monitoring disease progression. While Deep Learning (DL) architectures, such as U-Net and its extensions are very effective in medical image segmentation tasks, they often struggle with preserving fine-grained details and global contextual information. This is especially challenging for MRI data segmentation, where anatomical structures are characterized by irregular boundaries and variations in shape, contrast, and scale. To address this challenge, we propose a novel DL architecture for MRI segmentation across different anatomical structures. Specifically, the architecture introduces a module, named Multi-Resolution Feature Fusion (MRFF), that can be easily integrated into any U-Net-like architecture. The MRFF is integrated in all levels of an encode-decoder structure, along with attention mechanisms and skip connections to extract features at multiple resolutions, enabling the model to capture both fine-grained details and global contextual information. We evaluate the MRFFU-Net on two publicly available benchmark MRI datasets of different anatomical targets; one for Cerebrospinal Fluid (CSF) segmentation in spinal MR scans, and one for left atrium cardiac segmentation, from the Medical Segmentation Decathlon (MSD) challenge. Experimental results indicate that MRFFU-Net outperforms state-of-the-art models across multiple evaluation metrics, demonstrating its effectiveness in MRI segmentation.**

## I. INTRODUCTION

Clinical assessment and diagnosis of diseases often relies on medical imaging tools, such as Magnetic Resonance Imaging (MRI). The segmentation of regions of interest in MRI scans, such as the myocardium or Cerebrospinal Fluid (CSF), is a crucial process in detecting and monitoring disease progression. However, the analysis and curation of MRI scans involves manual delineation of anatomical structures, which is a costly and time-consuming process [1]. For example, CSF segmentation which is crucial for diagnosing a range of neurological conditions such as Alzheimer's disease, is considered challenging because CSF is characterized by low-contrast boundaries and high similarity with the surrounding structures. Therefore, these challenges have created a growing demand for automated segmentation techniques that can be used to assist clinicians in this process.

Recent advances in Deep Learning (DL) techniques have been developed for the segmentation of anatomical structures from MRI scans, such as brain tumors [2] and cardiac diseases [3]. For example, some methods have explored feature aggregation by combining the output of multiple Convolutional Neural Networks (CNNs) [4], while others have proposed corrective diffusion models for brain MRI segmentation by correcting systematic errors [5]. Some methods have also been developed that offer interpretations of their segmentation results [6]. Other approaches have been proposed in the literature for MRI segmentation, using the U-Net architecture and its extensions [7], including variations of the Attention U-Net [8], U-Net 3+ [9], U-Net++ [10], Dense-Inception U-Net [11] and 3D U-Net [12], among others. Similarly, additional convolutional operations were introduced in MultiResUNet [13] to enhance the semantic alignment from the encoder to the decoder part of the architecture. These methods introduce additional components to improve the localization performance of the model, such as skip connections [9], residual connections [14] and attention mechanisms [15], to capture the boundary irregularities and inhomogeneity that characterize the anatomical structures of interest. However, these methods fuse features using static operations like concatenation or addition, without considering learning features across multiple resolutions to capture both fine grained details of anatomical structures and global contextual information.

To address this challenge, some of the existing DL works in the literature have proposed components in their architectures to perform feature fusion for MRI segmentation tasks, such as spinal fracture [16] and nasopharyngeal [17] segmentation. For example, MMNet [18], performs left ventricular segmentation by applying dilated convolutions in the encoder to obtain information from different scales and uses skip connections in the decoder part of the architecture. However, these methods typically focus on a single type of anatomical structure and lack evaluation across anatomically diverse datasets. Other methods propose hybrid models that combine CNNs with transformer architectures, such as HM-Net [19] that proposed a hybrid pyramid attention module that integrates cross-over attention and residual modules between the decoder and encoder parts. Similarly, DCSwin-UNet [20] was proposed as a hybrid dual encoder U-Net architecture that incorporates both CNN and Swin Transformer encoders under a unified framework for brain tumor segmentation. However, these hybrid methods that combine transformer-based models typically present increased model complexity that introduces additional computational overhead and requires large-scale datasets for model training. Furthermore, these methods perform feature fusion at specific stages of the architecture, for instance only in the outer layers of the decoder, or they do not integrate information from multiple different feature resolutions, which can hinder their capacity of capturing fine and coarse features.

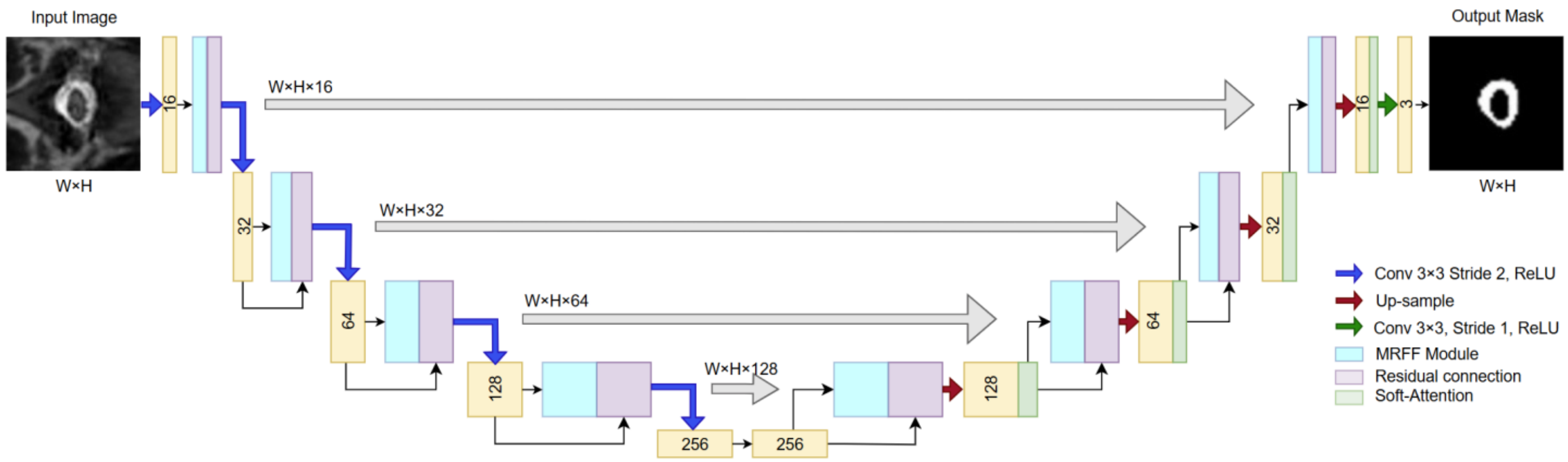

Fig. 1. The overall MRFFU-Net architecture.

In this work we propose a novel U-Net-based architecture for MRI segmentation across different anatomical structures. Specifically, we introduce a module, named Multi-Resolution Feature Fusion (MRFF), that enables the network to capture both fine-grained details and global contextual information, which is critical for segmenting MRI scans considering the structural irregularities present in anatomical regions of interest. The proposed architecture employs an encoder-decoder structure with skip connections and attention mechanisms to preserve spatial information enable precise localization of medical structures. The MRFF modules are integrated both in the encoder and decoder part of the architecture to extract features that capture the structural variations at different receptive fields, which are then aggregated via pointwise convolution to preserve spatial consistency. We validate the performance of the proposed architecture on two benchmark datasets for MRI segmentation, including CSF segmentation, as well as left atrium segmentation of cardiac MRI data.

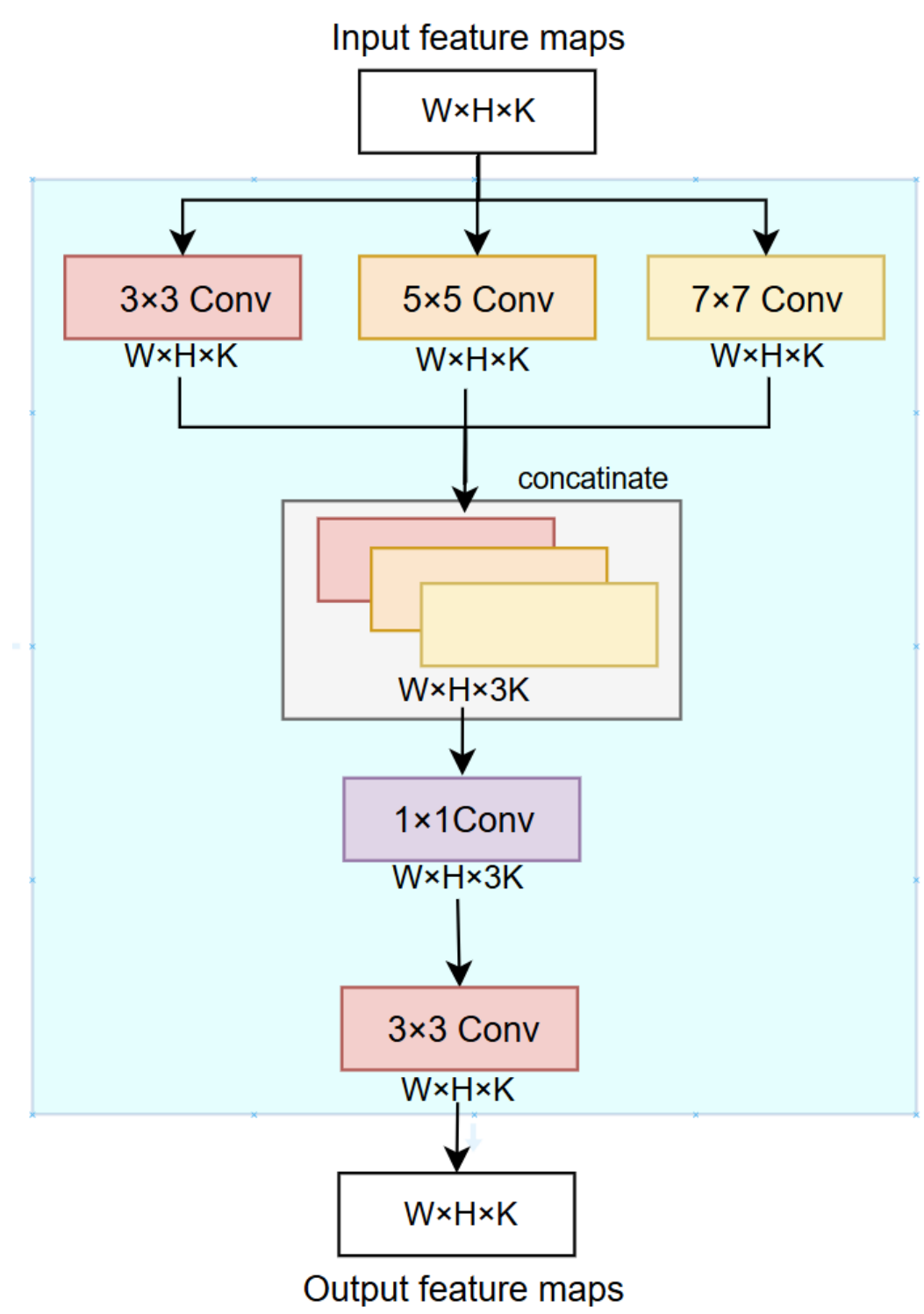

Fig. 2. The proposed MRFF module.

## II. METHODOLOGY

Figure 1 illustrates the architecture of a U-Net, enhanced with the proposed MRFF modules, uniquely combined with soft attention modules and skip connections to facilitate contextual learning and spatial localization. The MRFF modules and

### A. *Multi-Resolution Feature Fusion Module*

The MRFF module is illustrated in Fig. 2. It is designed to extract feature maps of different resolutions from the network, to capture fine grained and contextual information of the anatomical structures of interest, which are typically characterized by boundary irregularities. An MRFF module extracts a set of features of multiple receptive fields that can be derived by applying filters of three different kernel sizes, 3×3, 5×5 and 7×7 to input feature maps. This set of features is passed through a 1×1 convolution layer in order to fuse the contextual and global information from different layers of the network and control the channel dimensionality. This can be particularly useful in MRI data, where structures of interest vary in size across slices of in-patient scans. A convolution operation with a 3×3 kernel is then performed to derive a refined feature volume that is the output of the MRFF module of the same spatial dimension as the input feature maps it received.

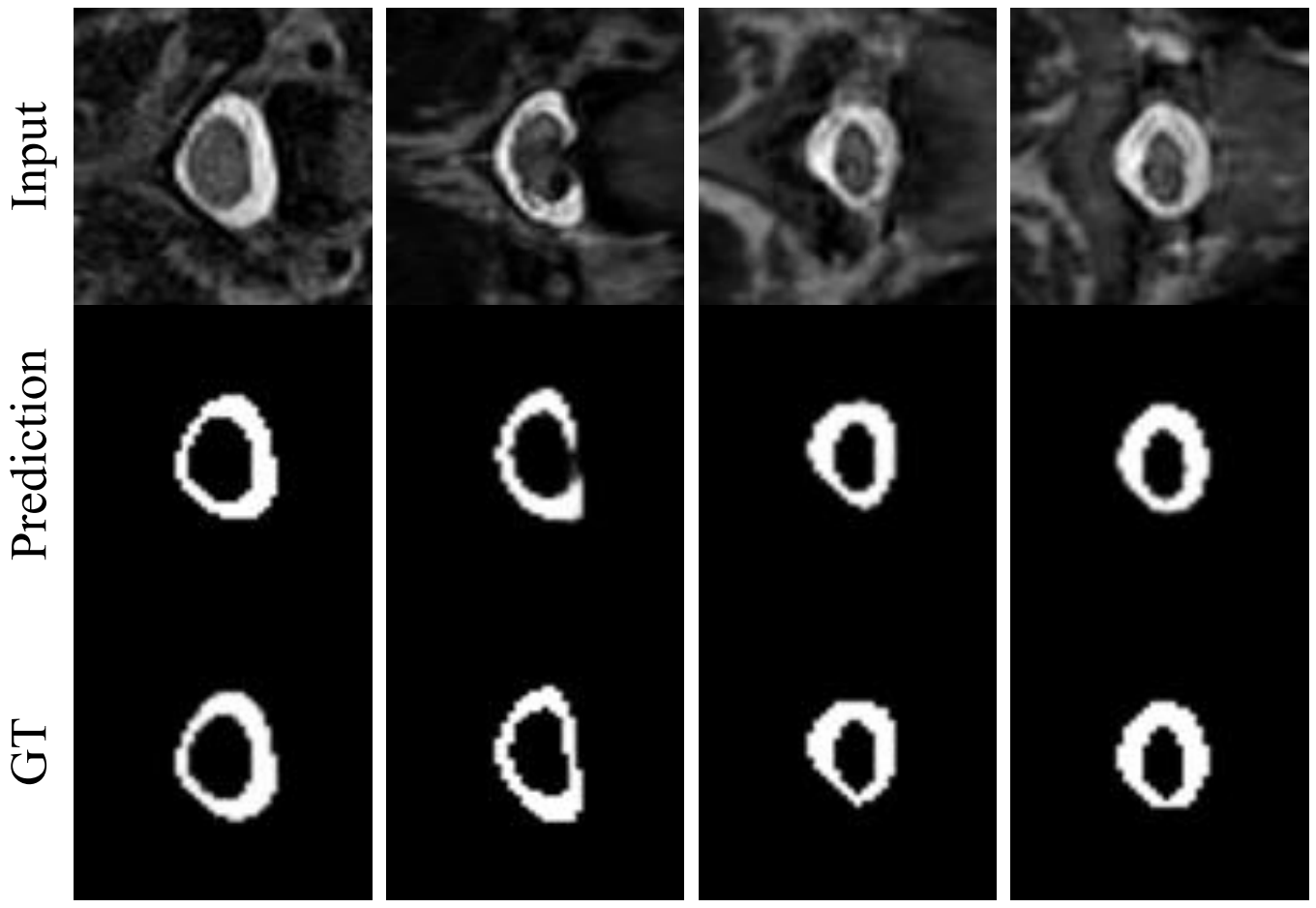

Fig. 3. Qualitative results on the spinal cord MRI dataset for CSF segmentation, illustrating the input image, predicted CSF segmentation mask of MRFFU-Net and ground truth (GT) image.

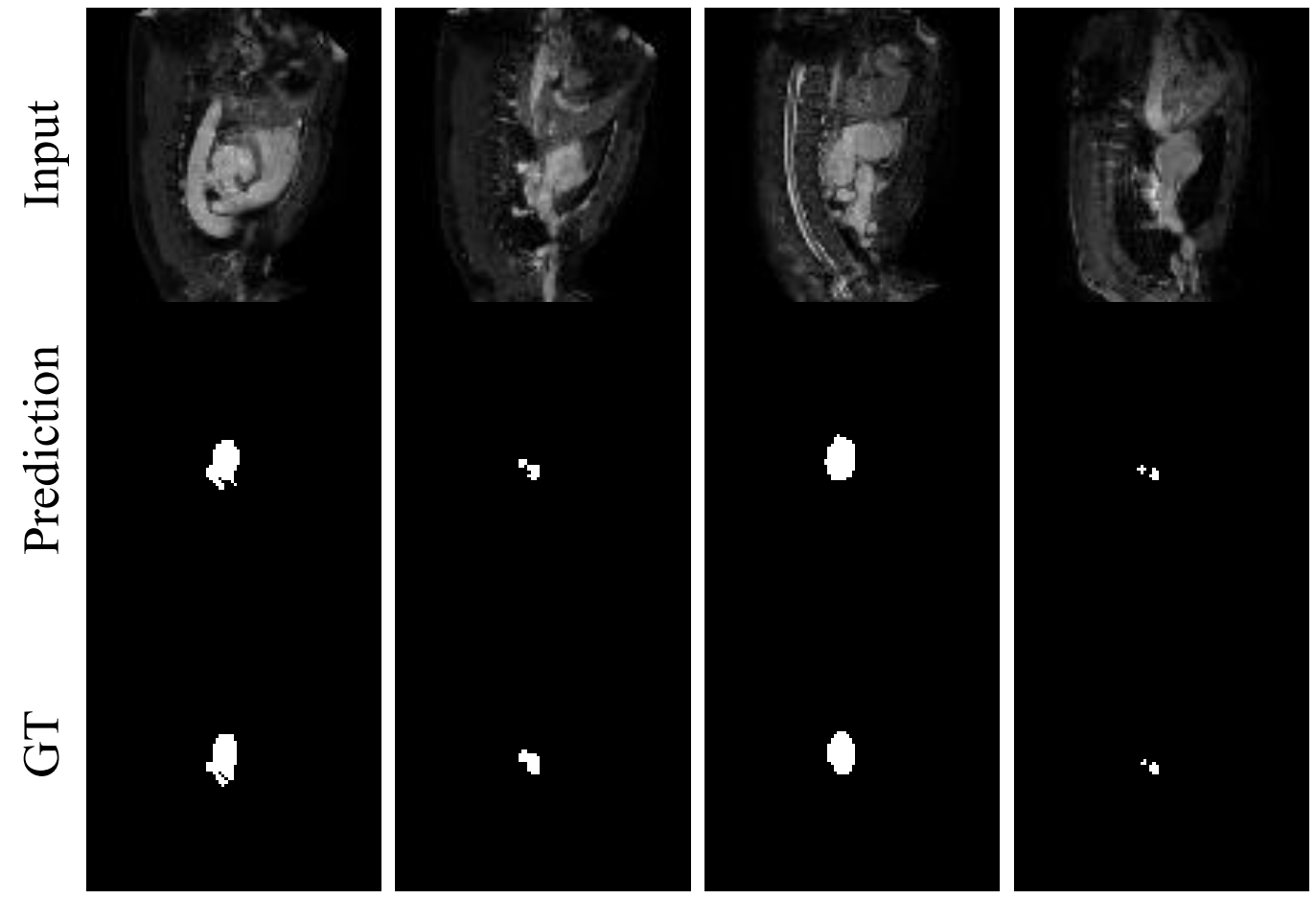

Fig. 4. Qualitative results on the cardiac MRI dataset for left atrium segmentation, illustrating the input image, predicted mask of MRFFU-Net and ground truth (GT) image.

### *B. Model Architecture*

The proposed architecture adopts a U-shaped structure, which is commonly used in medical image segmentation tasks, that is composed of a contracting path that involves an encoder and an expansion path of a decoder. The encoder path is composed of a series of convolutional blocks that progressively reduce spatial dimensions while increasing filter depth, capturing fine-to-coarse features, while mitigating information loss. Each convolutional block consists of a convolutional layer that has stride equal to 2, followed by a ReLU activation function and a batch normalization operation. Importantly, each convolutional block in the encoder is followed by an MRFF module, which extracts features at multiple resolutions (small, medium, and large) using convolutional kernels of varying sizes to enhance feature representations with contextual information before downsampling. The encoder consists of a total of four MRFF blocks, and the output of the last one is followed by a convolutional layer of a number of 256 feature map channels. Skip residual connections are employed at each level of the network, from the encoder to the decoder, to facilitate gradient flow and alleviate the problem of vanishing gradient [21].

The decoder path of the architecture incorporates a series of transposed convolutional blocks, with soft-attention modules to up-sample the feature maps and guide the network to focus on relevant features to perform precise localization of the regions of interest. The use of soft-attention modules is important as it enables the dynamic feature selection of fine-grained details in the derived feature maps while restoring spatial resolution. Skip residual connections from the encoder are concatenated with the corresponding up-sampled feature maps of the decoder to preserve fine-grained details. Similar to the encoder, MRFF modules are integrated into the decoder to enhance feature representation at multiple resolutions by performing feature fusion at each level of reconstruction. The final layer of the network uses a convolution block with stride equal to 1 to produce the pixel-wise segmentation map with the same spatial dimension as the input image.

Overall, the proposed method introduces a novel synergic integration of multi-resolution feature fusion and soft attention localization, in a unified framework for MRI segmentation. Unlike existing methods that incorporate attention mechanisms or feature fusion in their architecture, the proposed architecture integrates MRFF modules at both the encoder and decoder paths, enabling consistent enhanced feature learning across all network layers. This allows the model to adaptively capture fine structural variations and contextual patterns, which are crucial in MRI data where anatomical regions vary in size and have irregular boundaries. Additionally, the use of pointwise feature fusion and residual connections in the network preserve low-level details and enable gradient flow during training, especially in deeper layers of the network.

### *C. Optimization Objective*

The optimization objective of the proposed methodology employs the well-known Binary Cross-Entropy (BCE) loss function, which is widely used for pixel-level binary classification tasks and is suitable for the segmentation of MRI images. BCE quantifies the difference between the predicted output of the model and the ground truth mask and is defined as:

$$L_{BCE}(y,\hat{y}) = -\frac{1}{N}\sum_{i=1}^{N} y_i \log \hat{y}_i + (1-y_i)\log(1-\hat{y}_i) \quad (1)$$

where $N$ is the number of pixels, $y_i \in \{0,1\}$ is the ground truth label for pixel $i$ and $\hat{y}_i$ is the predicted label of that pixel derived by the model.

## III. EXPERIMENTS

### A. Experimental Setup

The segmentation performance of the proposed methodology was evaluated on two publicly available MRI datasets with distinct anatomical structures. The first dataset [22] contains MRI scans of the spinal cord, from which a subset was selected that consists of T2-weighted (T2w) sequences from multiple patients and MRI scanners, with expert CSF annotations. We followed the post-processing steps described in [22] to retrieve 2,116 T2w axial MR slices from the available anatomical scans across 10 distinct patients, along with corresponding CSF annotations. The second dataset was obtained from the Medical Segmentation Decathlon (MSD) [23] challenge that contains 10 distinct segmentation tasks across various anatomical targets and modalities. Specifically, we used the Heart dataset for the task of segmenting the left atrium, which was acquired as part of the 2013 Left Atrial Segmentation Challenge. This dataset consists of 30 mono-modal MRI scans, with 20 cases provided for training with available expert annotations. This dataset was selected due to the large anatomical variability.

To evaluate the performance of the proposed architecture, we employed four widely used metrics that include Mean Absolute Error (MAE), Mean Squared Error (MSE), Intersection over Union (IoU), and the Dice coefficient [24]. The comparative evaluation of the proposed methodology on the benchmark MRI datasets involved state-of-the-art methods that employ different network architectures, including ArachNet [6] and various U-Net architectures, including vanilla U-Net [25], Attention U-Net [26], U-Net++ [27] and U-Net 3+ [28].

The proposed methodology is implemented in Python 3.8 using the TensorFlow 2.11 framework on an NVIDIA RTX 3080 GPU with 24GB memory. The Adam optimizer was used with a batch size of 32 with a learning rate of 0.001 and early stopping was employed based on the validation loss to prevent overfitting. The size of the input images was set to 224×224 pixels and data augmentation included random horizontal and vertical flips. The experiments for the spinal cord dataset were conducted using a 5-fold validation, with 8 patients used for training and 2 for testing at each fold, whereas for the Heart dataset we employed a standard 80% training, 20% testing and 10% validation split of the scans with available mask annotations.

### B. Experimental Results

The quantitative results of the proposed methodology for the CSF segmentation task are summarized in Table I, against the comparative methods. It can be observed that MRFFU-Net outperforms all other methods in terms of Dice coefficient (0.886) and IoU (0.843), demonstrating its ability to capture the CSF regions accurately and maintain spatial consistency. Also, the proposed method achieves the lowest MAE (0.011) and MSE (0.005) scores, indicating significantly reduced pixel-wise errors compared to the state-of-the-art methods, which highlights the capacity of MRFFU-Net to produce accurate segmentation maps with respect to the ground truth. Table II reports the performance of the proposed method on the Heart dataset from the MSD challenge in terms of MAE, MSE, Dice and IoU scores against the comparative U-Net architectures. The proposed MRFFU-Net architecture achieves the highest overall performance across all metrics, reaching a Dice score of 0.879 and an IoU score of 0.816 respectively, outperforming all other comparative networks. These results demonstrate the capacity of the proposed architecture in segmenting various anatomical structures in MRI scans.

Figure 3 illustrates the qualitative results of the proposed methodology to further demonstrate the capacity of MRFFU-Net in the context of CSF segmentation. It can be observed that MRFFU-Net effectively segments the CSF regions, capturing the anatomical structures while maintaining consistency with the ground truth masks. Notably, even in cases where the CSF regions demonstrate morphological variations, MRFFU-Net accurately localizes the target regions. Some minor deviations can be observed in boundary regions, which could be attributed to partial volume effects or inherent imaging noise. Similarly, the qualitative results of MRFFU-Net on the MSD Heart dataset are presented in Figure 4, where it can be observed that the predicted segmentation masks consistently localize the left atrium region of interest, even in cases of high variability, which indicates the capacity of the proposed network to capture small and variable structures.

TABLE I. RESULTS IN TERMS OF MAE, MSE, DICE SCORE AND IOU AGAINST COMPARATIVE METHODS ON THE SPINAL CORD CSF DATASET.

| Method | Metric | | | |
|---|---|---|---|---|
| | ***MAE↓*** | ***MSE↓*** | ***Dice↑*** | ***IoU↑*** |
| *U-Net* [25] | 0.031 | 0.021 | 0.786 | 0.687 |
| *U-Net++* [27] | 0.052 | 0.024 | 0.733 | 0.658 |
| *Attention U-Net* [26] | 0.042 | 0.023 | 0.633 | 0.650 |
| *U-Net 3+* [28] | 0.032 | 0.022 | 0.797 | 0.697 |
| *ArachNet* [6] | 0.017 | **0.005** | 0.833 | 0.717 |
| *MRFFU-Net* | **0.011** | **0.005** | **0.886** | **0.843** |

TABLE II. RESULTS IN TERMS OF MAE, MSE, DICE SCORE AND IOU AGAINST COMPARATIVE METHODS ON THE MSD HEART DATASET.

| Method | Metric | | | |
|---|---|---|---|---|
| | ***MAE↓*** | ***MSE↓*** | ***Dice↑*** | ***IoU↑*** |
| *U-Net* [25] | 0.007 | 0.005 | 0.859 | 0.783 |
| *U-Net++* [27] | 0.009 | 0.011 | 0.739 | 0.659 |
| *Attention U-Net* [26] | 0.003 | 0.002 | 0.864 | 0.801 |
| *U-Net 3+* [28] | 0.002 | **0.001** | 0.866 | 0.799 |
| *MRFFU-Net* | **0.001** | **0.001** | **0.879** | **0.816** |

## IV. DISCUSSION AND CONCLUSIONS

In this study, we propose a novel DL architecture, namely MRFFU-Net, for MRI segmentation across different anatomical regions. The architecture introduces the MRFF module that enables the network to effectively capture both fine-grained details and global information, by performing feature fusion across different resolutions. We have integrated MRFFs at all levels of the network along with soft attention mechanisms and skip residual connections to refine feature representations and enable information flow throughout the network. This makes MRFFU-Net adept at handling the inconsistent boundaries and intra-class structure variations commonly present in MRI scans. Experimental results on two MRI benchmark datasets for cardiac and CSF segmentation demonstrated that MRFFU-Net outperforms the state-of-the-art methods across various evaluation metrics. In future work, we aim to extend the evaluation of MRFFU-Net to a broader range of medical datasets and assess its capacity for segmentation across different domains. Furthermore, we will explore the incorporation of domain-adaptation techniques and semi-supervised learning to assess the performance of MRFFU-Net in scenarios with limited annotated data.

## ACKNOWLEDGMENT

This work was supported by the European Union SoftReach project under Grant No. 101099145. (https://softreach.eu/)

## REFERENCES

[1] C. Petitjean and J.-N. Dacher, “A review of segmentation methods in short axis cardiac MR images,” *Medical image analysis*, vol. 15, no. 2, pp. 169–184, 2011.

[2] D. Ma, H. Zhang, and L. Wang, “Deep learning methods and applications in brain imaging for the diagnosis of neurological and psychiatric disorders,” *Frontiers in Neuroscience*, vol. 18. Frontiers Media SA, p. 1497417, 2024.

[3] C. Chen, C. Qin, H. Qiu, G. Tarroni, J. Duan, W. Bai, and D. Rueckert, “Deep learning for cardiac image segmentation: a review,” *Frontiers in cardiovascular medicine*, vol. 7, p. 25, 2020.

[4] W. Chen, X. Tan, J. Zhang, G. Du, Q. Fu, and H. Jiang, “A robust approach for multi-type classification of brain tumor using deep feature fusion,” *Frontiers in Neuroscience*, vol. 18, p. 1288274, 2024.

[5] W. Li, W. Huang, and Y. Zheng, “CorrDiff: Corrective diffusion model for accurate MRI brain tumor segmentation,” *IEEE Journal of Biomedical and Health Informatics*, vol. 28, no. 3, pp. 1587–1598, 2024.

[6] G. Dimas, P. G. Kalozoumis, P. Vartholomeos, and D. K. Iakovidis, “ArachNet: Interpretable Sub-Arachnoid Space Segmentation Using an Additive Convolutional Neural Network,” in *2024 IEEE International Symposium on Biomedical Imaging (ISBI)*, 2024, pp. 1–5.

[7] N. Siddique, S. Paheding, C. P. Elkin, and V. Devabhaktuni, “U-net and its variants for medical image segmentation: A review of theory and applications,” *IEEE access*, vol. 9, pp. 82031–82057, 2021.

[8] S. Wang, Z. Jiang, H. Yang, X. Li, and Z. Yang, “Automatic segmentation of lumbar spine MRI images based on improved attention U-net,” *Computational Intelligence and Neuroscience*, vol. 2022, no. 1, p. 4259471, 2022.

[9] A. A. Sifar, M. Shamsi, and M. H. Sedaaghi, “A modified U-Net 3+ based framework for automated segmentation of ventricle regions in brain MRI images,” *Signal, Image and Video Processing*, vol. 19, no. 7, pp. 1–13, 2025.

[10] S. A. Mortazavi-Zadeh, A. Amini, and H. Soltanian-Zadeh, “Brain tumor segmentation using U-net and U-net++ networks,” in *2022 30th International Conference on Electrical Engineering (ICEE)*, 2022, pp. 841–845.

[11] Z. Zhang, C. Wu, S. Coleman, and D. Kerr, “DENSE-INception U-net for medical image segmentation,” *Computer methods and programs in biomedicine*, vol. 192, p. 105395, 2020.

[12] W. Chen, B. Liu, S. Peng, J. Sun, and X. Qiao, “S3D-UNet: separable 3D U-Net for brain tumor segmentation,” in *International MICCAI Brainlesion Workshop*, 2018, pp. 358–368.

[13] N. Ibtehaz and M. S. Rahman, “MultiResUNet: Rethinking the U-Net architecture for multimodal biomedical image segmentation,” *Neural networks*, vol. 121, pp. 74–87, 2020.

[14] Y.-Z. Li, Y. Wang, Y.-H. Huang, P. Xiang, W.-X. Liu, Q.-Q. Lai, Y.-Y. Gao, M.-S. Xu, and Y.-F. Guo, “RSU-Net: U-net based on residual and self-attention mechanism in the segmentation of cardiac magnetic resonance images,” *Computer Methods and Programs in Biomedicine*, vol. 231, p. 107437, 2023.

[15] Z. Ji, X. Han, T. Lin, and W. Wang, “A dense-gated U-Net for brain lesion segmentation,” in *2020 IEEE International Conference on Visual Communications and Image Processing (VCIP)*, 2020, pp. 104–107.

[16] M. U. Saeed, W. Bin, J. Sheng, H. M. Albarakati, and A. Dastgir, “MSFF: An automated multi-scale feature fusion deep learning model for spine fracture segmentation using MRI,” *Biomedical Signal Processing and Control*, vol. 91, p. 105943, 2024.

[17] Y. Hao, H. Jiang, Z. Diao, T. Shi, L. Liu, H. Li, and W. Zhang, “MSU-Net: Multi-scale Sensitive U-Net based on pixel-edge-region level collaborative loss for nasopharyngeal MRI segmentation,” *Computers in Biology and Medicine*, vol. 159, p. 106956, 2023.

[18] Z. Wang, Y. Peng, D. Li, Y. Guo, and B. Zhang, “MMNet: A multi-scale deep learning network for the left ventricular segmentation of cardiac MRI images,” *Applied Intelligence*, vol. 52, no. 5, pp. 5225–5240, 2022.

[19] G. Zhao, X. Zhu, X. Wang, and F. Yan, “HM-Net: Hybrid multi-scale cross-order fusion network for medical image segmentation,” *Biomedical Signal Processing and Control*, vol. 98, p. 106658, 2024.

[20] R. Rekha, P. Shruti, M. Deekshitha, and J. Akash, “DCSwin-UNet: Dual Encoder U-Net based on CNN and Swin Transformer with Trainable Multiplication Layer for brain tumor segmentation from MRI images,” *Biomedical Signal Processing and Control*, vol. 110, p. 108325, 2025.

[21] K. He, X. Zhang, S. Ren, and J. Sun, “Deep residual learning for image recognition,” in *Proceedings of the IEEE conference on computer vision and pattern recognition*, 2016, pp. 770–778.

[22] J. Cohen-Adad, E. Alonso-Ortiz, M. Abramovic, C. Arneitz, N. Atcheson, L. Barlow, R. L. Barry, M. Barth, M. Battiston, C. Büchel, and others, “Open-access quantitative MRI data of the spinal cord and reproducibility across participants, sites and manufacturers,” *Scientific data*, vol. 8, no. 1, p. 219, 2021.

[23] M. Antonelli, A. Reinke, S. Bakas, K. Farahani, A. Kopp-Schneider, B. A. Landman, G. Litjens, B. Menze, O. Ronneberger, R. M. Summers, and others, “The medical segmentation decathlon,” *Nature communications*, vol. 13, no. 1, p. 4128, 2022.

[24] D. Müller, I. Soto-Rey, and F. Kramer, “Towards a guideline for evaluation metrics in medical image segmentation,” *BMC Research Notes*, vol. 15, no. 1, p. 210, 2022.

[25] O. Ronneberger, P. Fischer, and T. Brox, “U-net: Convolutional networks for biomedical image segmentation,” in *Medical image computing and computer-assisted intervention–MICCAI 2015: 18th international conference, Munich, Germany, October 5-9, 2015, proceedings, part III 18*, 2015, pp. 234–241.

[26] O. Oktay, J. Schlemper, L. L. Folgoc, M. Lee, M. Heinrich, K. Misawa, K. Mori, S. McDonagh, N. Y. Hammerla, B. Kainz, and others, “Attention u-net: Learning where to look for the pancreas,” *arXiv preprint arXiv:1804.03999*, 2018.

[27] Z. Zhou, M. M. Rahman Siddiquee, N. Tajbakhsh, and J. Liang, “Unet++: A nested u-net architecture for medical image segmentation,” in *International workshop on deep learning in medical image analysis*, 2018, pp. 3–11.

[28] H. Huang, L. Lin, R. Tong, H. Hu, Q. Zhang, Y. Iwamoto, X. Han, Y.-W. Chen, and J. Wu, “Unet 3+: A full-scale connected unet for medical image segmentation,” in *ICASSP 2020-2020 IEEE international conference on acoustics, speech and signal processing (ICASSP)*, 2020, pp. 1055–1059.